\documentclass[final]{cvpr}

\usepackage{times}
\usepackage{epsfig}
\usepackage{graphicx}
\usepackage{amsmath}
\usepackage{amssymb}

\usepackage[pagebackref=true,breaklinks=true,colorlinks,bookmarks=false]{hyperref}

\def\cvprPaperID{****} 
\def\confYear{CVPR 2021}

\begin{document}

\title{Multi-Granularity Network with Modal Attention for Dense Affective Understanding}
\author{Baoming Yan, Lin Wang, Ke Gao, Bo Gao, Xiao Liu, Chao Ban, Jiang Yang, Xiaobo Li \\
Alibaba Group\\
{\tt\small \{andy.ybm,youlin.wl,gaoke.gao,leo.gb,lemon.lx,banchao.bc,yangjiang.yj,xiaobo.lixb\}@alibaba-inc.com}






}

\maketitle

\begin{abstract}
 Video affective understanding, which aims to predict the evoked expressions by the video content, is desired for video creation and recommendation. In the recent EEV challenge, a dense affective understanding task is proposed and requires frame-level affective prediction.  In this paper, we propose a multi-granularity network with modal attention (MGN-MA), which employs multi-granularity features for better description of the target frame. Specifically, the multi-granularity features could be divided into frame-level, clips-level and video-level features, which corresponds to visual-salient content, semantic-context and video theme information. Then the modal attention fusion module is designed to fuse the multi-granularity features and emphasize more affection-relevant modals. Finally, the fused feature is fed into a Mixtures Of Experts (MOE) classifier to predict the expressions. Further employing model-ensemble post-processing, the proposed method achieves the correlation score of 0.02292 in the EEV challenge.
\end{abstract}

\section{Introduction}
Affective understanding is a task that aims to predict the viewer expressions evoked by the videos, which could be applied to video creation and recommendation. With the rapid increase of video content on the Internet, this technique becomes even more desired and has attracted enormous interest recently\cite{baveye2017affective,wang2015video}. However, due to the affective gap between video content and the emotional response, it remains a challenging topic. 

\begin{figure}[t]
\begin{center}
   \includegraphics[width=1\linewidth]{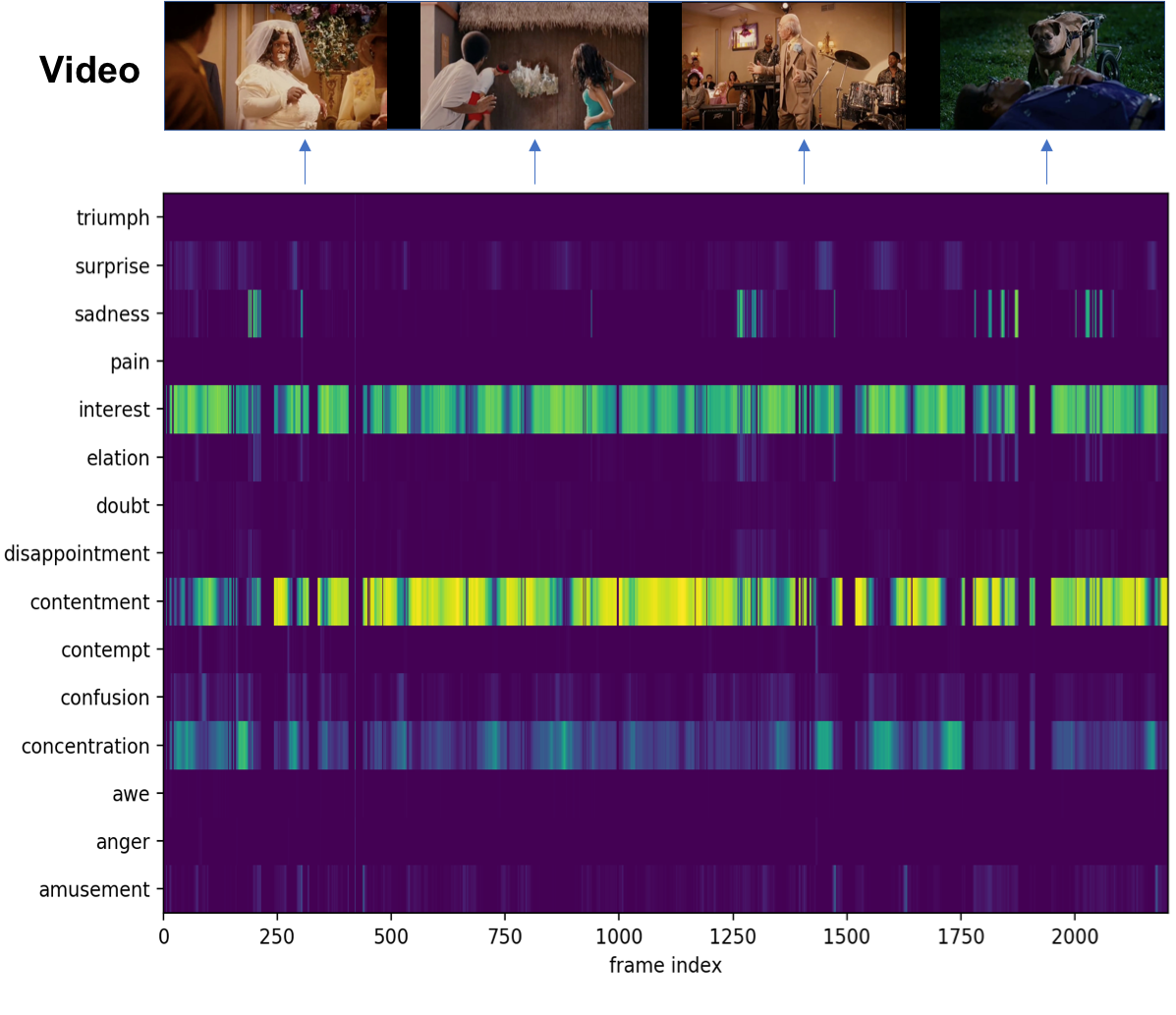}
\end{center}
   \caption{Visualization of scores of the 15 expressions scores for a video on EEV dataset.}
\label{fig:long}
\label{fig:onecol}
\end{figure}
The main methods in this area are multi-modal models, which combine visual and auditory features for affective predictions\cite{poria2015towards,sun2019gla}. However, for the recent proposed frame-level affective prediction task\cite{sun2021eev}, the expression of specific frame could be determined by signals of various timescales, these previous methods may be insufficient and imprecise. In this paper, we propose a Multi-Granularity Network with Modal Attention named MGN-MA, which extends previous multi-modal features into multi-granularity levels for frame-level affective understanding. Specifically, the multi-granularity features could be divided into three levels with various timescales, frame-level features, clips-level features and video-level features. Frame-level feature contains visual-salient content of the frame and audio style at the moment. Clips-level features represent semantic-context information around the frame, e.g. human behaviors, semantic information from speech. Video-level feature consists of the main affective information of the whole video, which is related to the theme of the video. Taken the above features into consideration, we employ a modal attention module with modal drop-out to further emphasize more affection-relevant modals. Finally, the representation goes through a MOE model\cite{jordan1994hierarchical} for the expression prediction. Experiments on the Evoked Expressions from Videos (EEV) dataset verify the effectiveness of our proposed method.
\begin{figure*}[htp]
\begin{center}
   \includegraphics[width=1\linewidth]{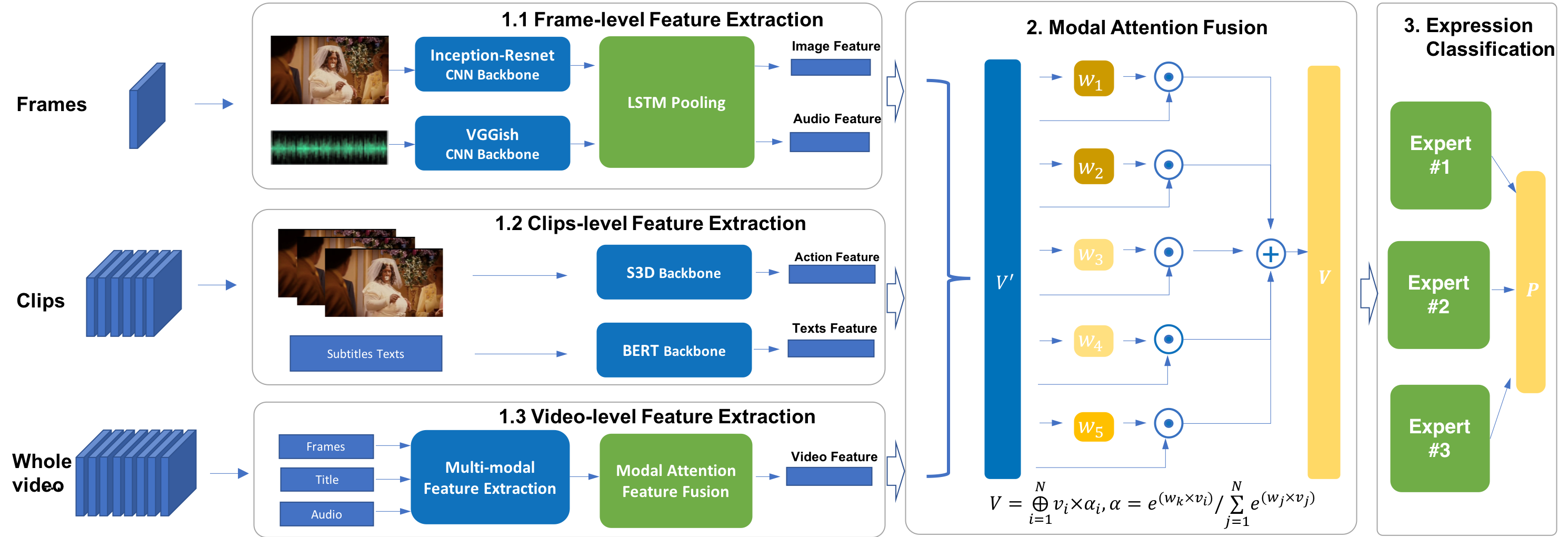}
\end{center}
   \caption{An overview of the proposed Multi-Granularity Network with Modal Attention (MGN-MA). The frame-level feature, the clips-level feature and video-level feature are extracted by multi-granularity feature construction module.Then the multi-granularity multi-modal feature is feed into modal attention fusion module and expression classification module.}
\label{fig2}
\end{figure*}
\section{Method}
The overall framework of our proposed Multi-Granularity Network with Modal Attention is illustrated in Fig.~\ref{fig2}, which could divided into Multi-Granularity Feature Construction, Modal Attention Fusion and Expression Classification. Taking frames, clips and the whole video as the multi-granularity input, three sub-networks are employed to extract frame-level, clips-level and video-level features and construct the Multi-Granularity Feature(MGF). Then the above MGF are fed into the Modal Attention Fusion(MAF) module to obtain the multi-modal feature. Finally, a MOE model is used to predict the evoked expressions of the frame.

\subsection{Multi-Granularity Feature Construction}
To provide sufficient information for frame-level expression prediction, we construct the Multi-Granularity Feature, which contains frame-level features, clips-level features and video-level features. More details of the feature extraction will be discussed as follows.

\subsubsection{Frame-level Features}
\,\,\,\,\,\,\,\textbf{Image Feature.}  We employ Inception-Resnet-v2\cite{szegedy2017inception} architecture pretrained on ImageNet\cite{deng2009imagenet} for image feature extraction, which could represent visual-semantic content appeared in the frame. We extract
the output in the last hidden layer as the feature and obtain a 1536-D vector for each frame. The image feature is performed at 6 Hz.

\textbf{Audio Feature.} We employ a VGG-style model provided by AudioSet\cite{gemmeke2017audio} trained on a preliminary version of YouTube8M\cite{abu2016youtube} for audio feature extraction. Following the method from \cite{hershey2017cnn}, the raw audio signal is first divided into 960 ms frames, then decomposed with a short-time Fourier transform applying 25 ms windows every 10 ms, which results the log-mel spectrogram patches of 96×64 bins. The spectrogram is further fed into the model and the 128-D embedding could be obtained.

\subsubsection{Clips-level Features}
\,\,\,\,\,\,\,\textbf{Action Feature.}  S3D\cite{xie2018rethinking} CNN backbone is used to extract Action feature for human behaviors. The S3D backbone is trained with MIL-NCE method\cite{miech2020end}, which employs self-supervised information between visual content and speech texts on HowTo100M\cite{miech2019howto100m} narrated video dataset. Around the target frame, we extract the context video clip with 32 frames sampled at 10 fps, and the output of the S3D Linear layer with 512-D is used as the embedding.

\textbf{Subtitle Feature.} Subtitles or automatically generated subtitles are downloaded for each video and the nearest speech text for each frame could be obtained. Then a base BERT model pretrained on the BooksCorpus (800M words)\cite{devlin2018bert} and English Wikipedia (2,500M words) is applied to extract the 768-D embedding of the speech text.

\subsubsection{Video-level Features}
To extract video-level feature, we trained a video-level expression prediction network on the EEV dataset\cite{sun2021eev} and the output of last hidden layer with 1024-D is used. The video-level expression is obtained by the averaged labels of frames. For each video, we take the 80 uniform selected video frames, audio frames and video title as input, and extract the multi-modal features by above Inception-Resnet-v2\cite{szegedy2017inception}, VGG-Style model\cite{hershey2017cnn} and BERT\cite{devlin2018bert} model. Features of video frames and audio frames are fed into its own NetVLAD sub-network for temporal pooling, and further merged into the video-level feature with the title feature. We propose a modal attention fusion module for feature fusion, which will be discussed in the following section.
\subsection{Modal Attention Fusion}
The modal attention fusion module is proposed to emphasize more affection-relevant features in the above multi-granularity multi-modal features. Specifically, The weighting value of each feature is adaptively predicted by an attention module $\alpha(\cdot)$, which outputs a normalized weight by softmax. The input of the modal attention fusion module is defined as the concatenation of above multi-granularity multi-modal features. By multiplying the weight to the corresponding feature, the total merged multi-modal feature $V$ could be obtained:
\begin{equation}
    V=\overset{N}{\underset{i=1}\bigoplus}{v_i\times \alpha_i}, \, \,  \alpha = e^{(w_k\times v_i)}/\sum_{j=1}^N{e^{(w_j\times v_j)}}
\end{equation}
where $\bigoplus$ is concatenation operator, $w_i$ is the weight of feature $v_i$, $N$ is total number of modals. A modal level dropout mechanism is performed by replacing specific feature by zeros vector randomly, which could improve the robustness of the model. 
\subsection{Expression Classification}
A 3-experts MOE\cite{jordan1994hierarchical} model is used for expression classification. To train all the parameters in our model, we minimize the classification loss between the predicted score and ground-truth score. The predicted score $o_{i,j}$ is first converted to probabilistic score by Sigmoid function:
\begin{equation}
    p_{i,j} = \frac{1}{1 + e^{-o_{i,j}}}
\end{equation}
Then the classification loss function is computed as :
\begin{equation}
    L =  -\frac{1}{B\times C}\sum_{i=1}^B\sum_{j=1}^Cy_{i,j}\log(p_{i,j})+(1-y_{i,j})\log(1-p_{i,j})
\end{equation}
where B is the batch size, C is the class number and $y_{i,j}$ is the $j_{th}$ expression label of the $i_{th}$ sample.
\section{Experiment}
\subsection{Datasets and Protocols}

\textbf{EEV dataset}  The dataset contains 8 million annotations of viewer facial reactions to 5,153 videos (370 hours). Each video is annotated at 6 Hz with 15 continuous evoked expression labels, corresponding to the facial expression of viewers who reacted to the video.

\textbf{Protocols} The performance is evaluated using correlation computed for each expression in each video, then averaged over the expressions and the videos. The correlation is based on scipy:
\begin{equation}
  r=\frac{\sum(x-m_x)(y-m_y)}{\sqrt{\sum(x-m_x)^2\sum(y-m_y)^2}}
\end{equation}
where \(x\) is the predicted values (0~1) for each expression, \(y\) is the ground truth value (0~1) for each expression, \(m_x\) is the average of \(x\) and \(m_y\) is the average of \(y\). Note that correlation is computed over each video.
\subsection{Implementation Details}
Our model is trained with Adam optimizer with 30 epochs. We start training with a learning rate of 0.0001. We use a mini-batch size of 1536. The dimension of the multi-modal feature is set to 1024. For the mixture of experts, we apply batch normalization before the non-linear layer. To avoid over-fitting, we select the snapshot of the model that gains the best result of correlation on the validation set. As some video might be deleted or made private, we employ the available 3024 videos for training and 730 videos for validation. The models are implemented in TensorFlow.
\subsection{Results}

To evaluate the importance of different granularity of features, we start with frame-level features and gradually add the clip-level features and video-level features, the experiment results on the test set of the EEV dataset are shown in Table 1. For the frame-level feature, the performance of image feature and audio feature are evaluated separately with the correlation score of 0.00349 and 0.00574, which implies that audio is more effective than visual content for affective prediction. The combination of image and audio feature could further increase the performance to 0.00739. After the introduction of clip-level features, the correlation score is remarkably increased by 0.0589. This indicates the importance of semantic-context information around the frame, which is not supplied from the image feature and audio feature. It is noted that the subtitle feature are more effective than action features, this means more semantic information exists in the subtitles. Moreover, the introduction of video-level feature further increases the performance by 0.00201, and proves the importance of the affective information from the whole video. The above results indicates that our proposed multi-granularity features are effective for affective prediction.
\begin{table}[htp]
    \centering
    \begin{tabular}{lc}
    \hline
    Input & Correction \\
    \hline
    Image     &  0.00349  \\
    Audio     &  0.00574  \\
    Image+Audio     &  0.00739  \\
    Image+Audio+Action    &   0.00893 \\
    Image+Audio+Action+Subtitle    &   0.01328 \\
    Image+Audio+Action+Subtitle+Video    &   0.01529 \\
    \hline
    \\
    \end{tabular}
    \caption{Comparison of different input modalities on EEV}
    \label{tab:my_label}
\end{table}

\begin{table}[htp]
    \centering
    \begin{tabular}{lc}
    \hline
    Method & Correction \\
    \hline
    MLP     &   0.01529  \\
    MOE     &  0.01718  \\
    MAF+MOE     &  0.01849  \\
    Ensemble    &   0.02292 \\
    \hline
    \\
    \end{tabular}
    \caption{Comparison of different model methods on EEV}
    \label{tab:my_label}
\end{table}
In addition to the proposed the multi-granularity feature, we further evaluate the performance of modal fusion module and expression classification module. As shown in Table 2, four methods are compared with the same multi-granularity feature. Compared with Multilayer Perception classification, MOE improves the correlation result by 0.00189. The model achieves the best performance when the number of expert is set as 3. When the Modal Attention Fusion (MAF) module is introduced to replace the feature concatenation fusion, the correlation score is improved from 0.01718 to 0.01849. We ensembles five models to further improve the performance, and  achieve a correlation score of 0.02292, which improves the result by 0.00443 compared with single model.
\section{Conclusion}
In this paper, the multi-granularity network with modal attention (MGN-MA) is proposed for dense affective predictions. MGN-MA consists of multi-granularity feature construction module, modal attention fusion module and expression classification module. The multi-granularity feature construction module could learn more semantic content and video theme information. The modal attention fusion module further improves the performance by emphasizing more affection-relevant features in the multi-granularity multi-modal features. The MGN-MA achives the correlation score of 0.02292 in the EEV challenge.

{\small
\bibliographystyle{ieee_fullname}
\bibliography{egbib}
}

\end{document}